\documentclass[10pt,twocolumn,letterpaper]{article}

\usepackage{iccv}
\usepackage{times}
\usepackage{epsfig}
\usepackage{graphicx}
\usepackage{amsmath}
\usepackage{amssymb}
\usepackage{url} 

\newcommand{\h}[1]{\hspace{#1}}

\newcommand{\argmin}{\mathop{\rm arg~min}\limits}

\usepackage[pagebackref=true,breaklinks=true,letterpaper=true,colorlinks,bookmarks=false]{hyperref}

\iccvfinalcopy 

\def\iccvPaperID{939} 

\ificcvfinal\fi
\begin{document}

\title{APAC: Augmented PAttern Classification with Neural Networks}

\author{Ikuro Sato\\
Denso IT Laboratory, Inc.\\
Tokyo, Japan\\
{\tt\small isato@d-itlab.co.jp}
\and
Hiroki Nishimura\\
Denso IT Laboratory, Inc.\\
Tokyo, Japan\\
{\tt\small hnishimura@d-itlab.co.jp}
\and
Kensuke Yokoi\\
DENSO CORPORATION\\
Aichi, Japan\\
{\tt\small kensuke\_yokoi@denso.co.jp}
}

\maketitle

\begin{abstract}
Deep neural networks have been exhibiting splendid accuracies in many of visual pattern classification problems.
Many of the state-of-the-art methods employ a technique known as data augmentation at the training stage.
This paper addresses an issue of decision rule for classifiers trained with augmented data.
Our method is named as APAC: the Augmented PAttern Classification, which is a way of classification using
the optimal decision rule for augmented data learning.
Discussion of methods of data augmentation is not our primary focus.
We show clear evidences that APAC gives far better generalization performance than 
the traditional way of class prediction in several experiments.
Our convolutional neural network model with APAC 
achieved a state-of-the-art accuracy on the MNIST dataset among non-ensemble classifiers.
Even our multilayer perceptron model beats some of the convolutional models with recently-invented
stochastic regularization techniques on the CIFAR-10 dataset.
\end{abstract}

\section{Introduction}

Output of an ideal pattern classifier satisfies two properties.
One is the invariance under replacement of a data point by another data point within the same class,
and we refer this as to intra-class invariance.
The other is the distinctiveness under replacement of a data point in one class
by a data point in another class,
and we refer this as to inter-class distinctiveness.
Good classifiers more or less have these properties for untrained data.

For a given class, there exists a set of transformations that leave the class label unchanged.
In case of visual object recognition of ``apple", the class label stays the same 
under different lighting conditions, backgrounds, and poses, to name a few.
One can expect that a classifier gains good intra-class invariance through learning dataset 
containing many images with these variations.

A classifier should also show inter-class distinctiveness to distinguish one class from the other.
If one construct a training dataset containing green apple class and red apple class, 
lighting condition must be paid careful attention, because important color feature may be spoiled 
under some lighting condition.
Appropriate types and ranges of variations depends on the problem setting.

For an image classifier to gain good intra-class invariance without compromising 
inter-class distinctiveness, there are largely two types of approaches.
One approach is to embed some mechanisms in classifiers to give robustness against intra-class variations.
One of the most successful classifiers would be Convolutional Neural Network (CNN) \cite{lecun-89e}.
CNN has two important building blocks: convolution and spatial pooling, which
give robustness against global and small local shifts, respectively.
These shifts are a typical form of intra-class variation.

Another approach is data augmentation, meaning that a given dataset is expanded by virtual means.
A common way is to deform original data in many ways using prior knowledge on intra-class variation.
Color processing and geometrical transformation (rotation, resizing, \etc) are typical operations
used in visual recognition problems.
Adding virtual data points amounts to making the points denser in the manifold that the class instances form.
Strong regularization effects are expected through augmented data learning.

Augmented data learning is also beneficial in an engineering point of view.
Dataset creation is a painstaking and costly part in product development.
Data augmentation allows the use of prior knowledge on recognition targets, 
which engineers do have in most cases,
and thus provides easy and cheep substitutes.
Secondly, quality of virtual data can be easily evaluated by human perception.
In case of visual recognition task, one can check virtual images whether they resemble real ones by eyes.

Many of state-of-the-art methods in generic object recognition problems use deep CNNs, trained on augmented datasets
comprising original data and deformed data 
(see recent works \cite{2015arXiv150102876W, 2014arXiv1409.4842S, NIPS2012_4824, 2015arXiv150201852H}).
It has been pointed out that CNN models with many layers
have great discriminative power; 
on the other hand, theoretical and methodological aspects of data augmentation are not fully revealed.

\subsection{Related work}

Data augmentation plays an essential role in boosting performance of generic object recognition.
Krizhevsky \etal used a few types of image processing, 
such as random cropping, horizontal reflection, and color processing, 
to create image patches for the ImageNet training \cite{NIPS2012_4824}.
More recently, Wu \etal vastly expanded the ImageNet dataset with many types of image processing 
including color casting, vignetting, rotation, aspect ratio change, and lens distortion on top of 
standard cropping and flipping~\cite{2015arXiv150102876W}.
Although these two works use different network architectures and computational hardware,
it is still interesting to see the difference in the performances levels.
The top-5 prediction error rate of the latter is 5.33\%, while that of the former is 16.42\%.
Such a large gap could be an implicit evidence that
richer data augmentation leads to better generalization.

Paulin \etal proposed a novel method for creating augmented datasets \cite{paulin:hal-00979464}.
It greedily selects types of transformations that maximize the classification performance.
The algorithm requires heavy computational resources, thus the exhaustive pursuit is almost intractable 
when deep networks with a huge number of parameters are trained.

Handwritten character/digit recognition has been an important problem for both industrial applications and 
algorithm benchmarking for a quarter century
\cite{conf/nips/YaegerLW96, lecun-98h, Simard03j.c.:best, lecun-89e, Ciresan:2010, Ciresan:2012b}.
The problem is relatively simple in a sense that there is no degree of freedom in the background
and that stroke can be easily modified.
Elastic distortion is a commonly used data augmentation technique that has a good property in 
giving a large degrees of freedom in the stroke forms, 
while leaving the topological structure invariant.
Indeed, data augmentation by elastic distortion is crucial in boosting classification performance
\cite{Simard03j.c.:best, Ciresan:2010, Ciresan:2012b}.

In case of pedestrian detection, use of synthetic pedestrians in real background \cite{6977491} 
and synthetic occlusion \cite{6728373} have been proposed.
Though these approaches give additional degrees of freedom in expanding training datasets,
we omit such means in this work.

Data augmentation can be categorized into two: off-line and on-line.
In this work, off-line data augmentation means to increase the number of data points by a fixed factor 
before the training starts.
The same instance is repeatedly used in the training stage until convergence~\cite{Simard03j.c.:best}.
On-line data augmentation means to increase the number of data points 
by creating new virtual samples 
at each iteration in the training stage (see representative works:~\cite{Ciresan:2012b, Ciresan:2010}).
There, random deformation parameters are sampled at each iteration, 
hence the classifier always ``sees" new samples during the training.
Cire{\c s}an \etal claims that on-line scheme greatly improves classification performance
because learning a very large number of samples likely avoids over-fitting~\cite{Ciresan:2012b, Ciresan:2010}.
Our work is mostly inspired by their work, and is focused on the on-line deformation.

Very recently, an website article reported a method named as Test-Time Augmentation~\cite{SanderDieleman},
where prediction is made by taking average of the output from many virtual samples,
though the algorithm is not fully described.

Tangent Prop \cite{560025} is a way to avoid over-fitting with implicit use of data augmentation.
Virtual samples are used to compute the regularization term that is defined as the sum of tangent distances,
each of which is the distance between an original sample and a slightly deformed one.
It is expected that classifier's output is stable in the vicinity of original data points,
but not necessarily so in other locations. 

\subsection{Contribution}

This paper proposes the optimal decision rule for a given data sample using classifiers trained with
augmented data.
We do not discuss methods of data deformation themselves.
Throughout this paper we assume that training is done with data samples deformed in on-line fashion.
That is, random deformation parameters are sampled at every iteration, and a deformed sample is used only once
and discarded after a single use.
Such training minimizes an expectation value of loss function over random deformation parameters.
We claim that class decision must be made so as to minimize the same expectation value for a given test sample.

We show by experiments that the proposed decision rule give lower classification error rates than 
the conventional decision rule.
APAC improves test error rate of CNN by 0.16\% for MNIST and by 9.72\% for CIFAR-10.
To the best of our knowledge, the improved error rate for MNIST is the best among non-ensemble classifiers reported
in the past.

Though we believe that the proposed decision rule is beneficial to any classification problem,
in which augmented data learning is applied, 
image classification problems are mainly discussed in this paper
because we have not conducted experiments in other fields.

\section{On-line data deformation}

On-line data deformation learning can generate classifiers with strong intra-class invariance.
Such learning generally consumes many iterations to reach a minimum of the objective function.
A vast number of training instances are processed because 
the number of instances increases linearly as the number of iterations increases.
In the on-line deformation scheme, 
the original data themselves are not trained explicitly --they are only trained probabilistically.

In this section we provide a formal definition of augmented data learning, 
which has been treated rather heuristically so far.
Let us first define the data deformation function as
$u: \mathbb{R}^{d} \rightarrow \mathbb{R}^{d}$, where $d$ is the dimension of the 
original data.\footnote{The data deformation function can be generalized to 
$u: \mathbb{R}^{d_0} \rightarrow \mathbb{R}^{d_1}$ with $d_0 \neq d_1$, 
but we consider $d_0 = d_1$ case in this study for simplicity.}
The function $u(x; \Theta)$ 
takes a datum $x \in \mathbb{R}^{d}$ and deformation-controlling parameters $\Theta = \{\theta_1, \cdots, \theta_K\}$,
and returns a virtual sample.
Each element of the set $\Theta$ is defined as a continuous random variable for convenience.
Some are responsible for continuous deformation; 
\eg, $\theta_1$ being scaling factor, $\theta_2$ being horizontal shift, \etc.
The other are responsible for discrete deformation;
\eg, if $\theta_3 \in [0, \frac{1}{2})$ horizontal side is flipped, 
and if $\theta_3 \in [\frac{1}{2}, 1]$ no side-flipping is performed, 
where $\theta_3 \sim {\cal U}(0,1)$.
We use class label $c$ in the superscript, $\Theta^c$, if deformation is done in a class-dependent fashion.
In this work, it is assumed that probability density functions of 
deformation parameters are given at the beginning and
held fixed during training and testing.
In the following, we consider two cases: 
1) the way of deformation being same for all classes, and 
2) the others.
We use the cross entropy as the loss function as it is most widely used for Deep Learning with supervised setting.
The cross entropy requires vector normalization in the output units, where we use the softmax function.

\subsection{Class-indistinctive deformation learning}

We first discuss the case 1). 
Let $i\in\{1,\cdots,N\}$ denote an index of original training data, 
$c_i\in\{1,\cdots,N_c\}$ denote the class index of $i$-th sample,
$W$ denote the set of all parameters to be optimized, and
$f(~\cdot~; W): \mathbb{R}^{d} \rightarrow \mathbb{R}_{>0}^{N_c}$ denote a function realized by a neural network with the softmax output units.
Let $f_c$ be the $c$-th component of the output, then
$\sum_{c=1}^{N_c} f_c = 1$ and $f_c > 0, \forall c \in \{1,\cdots, N_c\}$.
In the following, regularization terms are ignored for simplicity.
\newline
\newline
{\bf Class-indistinctive deformation learning:} \\
Given ${\cal D} = \{ (x_i, c_i) \}, i=1,\cdots,N$, find $W^\star$ such that
\begin{equation}
W^\star = \argmin_W J_{\cal D}(W),
\end{equation}
where the objective function $J_{\cal D}(W)$ is defined as
\begin{equation}
J_{\cal D}(W) = \sum_{i=1}^N \mathbb{E}_\Theta
\left[ - \ln \left( f_{c_i} \left( u \left( x_i; \Theta \right); W \right) \right) \right].
\label{eq-2}
\end{equation}

The expectation value is computed by marginalizing the cross entropy over 
deformation parameters that independently obey unconditional probability
densities $p_k(\theta_k), k=1, \cdots, K$.
By using appropriate random number generators, one can generate countlessly many virtual samples during training.
By sufficiently reducing the objective function, 
the classifier outputs a value close to the target value for an arbitrarily deformed training image.
That means, the classifier gains a high level of intra-class invariance with respect to the set of deformations applied,
without compromising inter-class distinctiveness.

Deformation must be meaningful for all classes in class-indistinctive deformation learning.
It may be homography transformation or global color processing such as gamma correction, to name a few.

A truly intra-class-invariant classifier would be obtained, if the integrals
$\mathbb{E}_\Theta [\cdot]  = \int \cdots \int \prod_k d\theta_k p_k(\theta_k) (\cdot)$
could be analytically calculated.
However, it is hard to integrate out in reality.
Then one needs to convert the integral into a sum of infinitely many terms,
\begin{equation}
\mathbb{E}_\Theta [~\cdot~] = 
\lim_{R\rightarrow\infty} \frac{1}{R} \sum_{\Theta = \Theta^{(1)}, \cdots, \Theta^{(R)}} (~\cdot~).
\end{equation}
Here, $\Theta^{(\ell)} = \{ \theta_1^{(\ell)}, \cdots, \theta_K^{(\ell)} \}$ is a set of deformation
parameters at $\ell$-th sampling, based on
the unconditional probability density functions $p_k(\cdot), k=1,\cdots,K$.
With this summation form, the objective function can be approximately minimized 
by widely-used mini-batch Stochastic Gradient Descent (SGD).
Note that a batch optimization algorithm is no longer applicable in a strict sense 
because the number of terms are infinite.
At each iteration in the optimization process, data indices and deformation parameters are randomly 
sampled to generate a mini-batch.
The mini-batch is discarded after a single use.
The total number of terms in the objective function is determined 
when the training is terminated.
It is clear from Eq.~(\ref{eq-2}) that the original data samples are not explicitly fed into the network.

We believe that the original data should not be used for validation,
as opposed to the statement made by Cire{\c s}an~\etal~\cite{Ciresan:2012b, Ciresan:2010},
where they claim that the original data can be used for validation.
The original and deformed data have strong correlations in the feature space 
especially when deformation is moderate.
Therefore, it is advised not to use the original training data to estimate the generalization performance.

In the experiment, we employ class-indistinctive deformation learning.

\subsection{Class-distinctive deformation learning}

Next, we discuss the case 2), where
the probability densities for deformation parameters depend on the classes.
Although such scheme requires one to design deformation in a
class-specific way, it is likely to give a stronger inter-class
distinctiveness to the classifier.
For example,
it is not probably a good idea to cast color with strong red component
to an image belonging to ``green apple" class, when
there is ``red apple" class, for an obvious reason.
But casting red color to an image belonging to, say, ``grape" class may be reasonable.
In the hand-written digit classification problem, 
Cire{\c s}an \etal have used different ranges of deformation
parameters for certain classes
\cite{Ciresan:2010}:
rotation and shearing applied to digit 1 and 7 are
less stronger than other digits.
Their work is another example of class-distinctive deformation learning.
\newline
\newline
{\bf Class-distinctive deformation learning:} \\
Given ${\cal D} = \{ (x_i, c_i) \}, i=1,\cdots,N$, find $W^\star$ such that
\begin{equation}
W^\star = \argmin_W \widetilde{J}_{\cal D}(W),
\end{equation}
where the objective function $\widetilde{J}_{\cal D}(W)$ is defined as
\begin{equation}
\widetilde{J}_{\cal D}(W) = \sum_{i=1}^N \mathbb{E}_\Theta
\left[ - \ln \left( f_{c_i} \left( u \left( x_i; 
\Theta \right); W \right) \right) | c_i \right].
\label{class-distinctive-cost}
\end{equation}

For an arbitrary $i$-th empirical sample,
expectation value is computed by marginalizing over 
deformation parameters:
$\mathbb{E}_{\Theta}[\cdot | c_i] = \int\cdots\int\prod_k d\theta_k 
p_k (\theta_k | c_i) (\cdot)$.
Here, the $k$-th deformation parameter 
obeys a class-conditional probability density $p_k (\theta_k | c_i)$.\footnote{There may be a case where 
certain types of deformation are only applied to selected class(es). In such a
case, a delta function is used as PDF to ``turn-off" the deformation for other classes; 
\ie, $p_k(\theta_k | c) = \delta (\theta_k)$.}

The optimization procedure is similar to that of the class-indistinctive
deformation learning, except that deformation parameters obey
conditional probabilities.
The integral can be rewritten by a sum of an infinite number of terms,
\begin{equation}
\mathbb{E}_{\Theta} \left[ ~\cdot ~| c \right]
= \lim_{R\rightarrow\infty} \frac{1}{R} \sum_{\Theta = \Theta^{c (1)}, \cdots, \Theta^{c(R)}} (~\cdot ~),
\end{equation}
where $\Theta^{c(\ell)} = \{\theta_1^{c(\ell)}, \cdots, \theta_K^{c(\ell)} \}$ is a
set of deformation parameters at $\ell$-th sampling, based on the 
PDFs: $p_k(\theta_k|c), k=1,\cdots,K, c=1,\cdots, N_c$.
Some form of SGD can be used to minimize the objective function
with a finite term approximation.

\section{Decision rule for augmented data learning}

In this section we propose a new way of classification, 
{\bf APAC: Augmented PAttern Classification}, and claim that
it gives the optimal class decision for augmented data learning
described in the previous section.
It is shown that 
a single feedforward of a given test sample is no longer optimal 
when one minimizes the expectation value at the training stage.
Cross entropy loss with softmax normalization is assumed
in the following discussion.
\newline
\newline
{\bf APAC for class-indistinctive deformation learning:} \\
Given parameters $W$ and data $x$, 
find $c^\star$ such that 
\begin{equation}
c^\star = \argmin_{c\in \{1,\cdots, N_c\}} J_{\{(x, c)\}}(W).
\label{APAC-indistinctive}
\end{equation}
\newline
{\bf APAC for class-distinctive deformation learning:} \\
Given parameters $W$ and data $x$, 
find $c^\star$ such that 
\begin{equation}
c^\star = \argmin_{c\in \{1,\cdots, N_c\}} \widetilde{J}_{\{(x, c)\}}(W).
\label{APAC-distinctive}
\end{equation}

\begin{figure}[t]
\begin{center}
\includegraphics[width=0.95\linewidth]{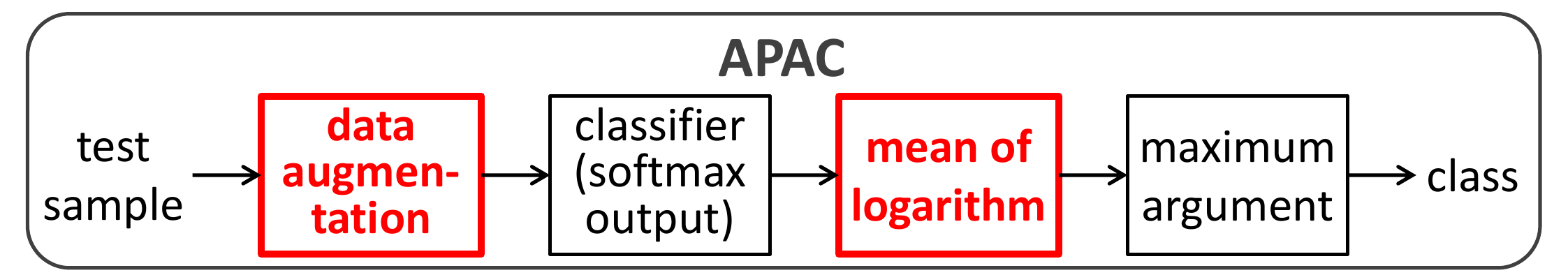}\\ \vspace{-0.0cm}
\includegraphics[width=0.95\linewidth]{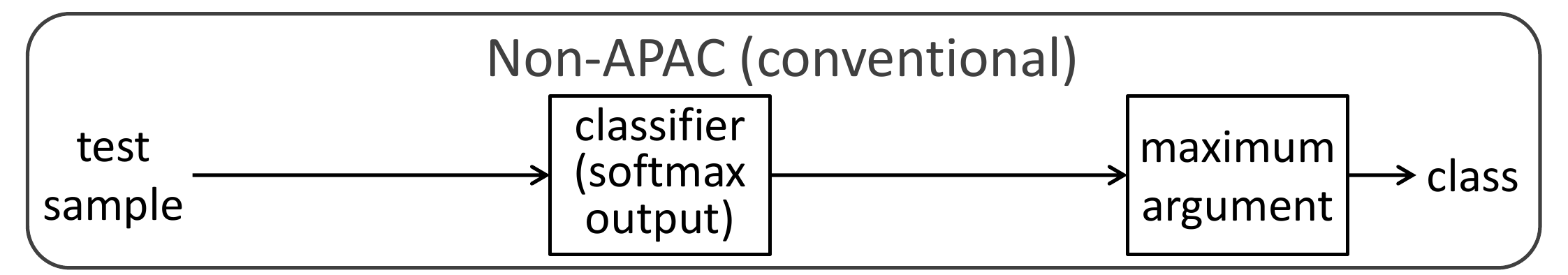}
\end{center}
\caption{APAC, the proposed way of classification (above).
Non-APAC, conventional way of classification (below).}
\label{fig-APAC}
\end{figure}

It is obvious from 
Eq.~(\ref{APAC-indistinctive}, \ref{APAC-distinctive})
that class decision making is an {\it optimization} process
requiring minimization of the expectation values.
The expectation value for a given data sample 
must be computed at test stage,
as it is minimized through training stage (with some approximation).
Note that the test sample itself is not fed into the classifier.
In practice, finite-term relaxation must be made at test stage
to estimate the expectation value:
\begin{eqnarray}
\mathbb{E}_\Theta [~\cdot~] &\simeq & \frac{1}{M}
\sum_{\Theta = \Theta^{(1)}, \cdots, \Theta^{(M)}} (~\cdot~)
\label{eq-finite-term-approximation}
\\
\mathbb{E}_\Theta [~\cdot~ | c] &\simeq & \frac{1}{M}
\sum_{\Theta = \Theta^{c(1)}, \cdots, \Theta^{c(M)}} (~\cdot~)
\end{eqnarray}
for the class-indistinctive case and class-distinctive case, respectively.
This means, a finite number of sets of deformation parameters must be randomly sampled
using the same probability density functions used in the 
training.
APAC requires to average the logarithms of the softmax output,
and then take the maximum argument to give optimal prediction.
The process flow is depicted in Fig.~\ref{fig-APAC}.
We emphasize that taking logarithm is an important step,
otherwise an irrelevant quantity gets minimized at the test stage and
classification performance likely degrades.
APAC is equivalent to picking the maximum argument of the product of the softmax output,
which is analogous to selecting the largest joint probability 
among individual class-probabilities of many virtual instances.
For a sufficiently trained classifier, it is expected that generalization performance asymptotically reaches the highest 
as the number of terms, $M$, increases.

The decision rule for class-distinctive deformation learning
requires to generate plural sets of virtual samples for a given test image.
Suppose one uses $N_d$ sets of 
deformations\footnote{$N_d=N_c$ when each class has a unique deformation set, and
$N_d<N_c$ when two or more classes share the same type of deformations.}
at the training stage, then
at the testing stage a data sample has to be deformed in $N_d M$ different ways.
Then average of $M$ logarithms of softmax output is computed 
for each class, using the corresponding deformation type.
A maximum argument is then picked to predict a class.


%
%

\section{Experiments}

Experiments on image classification are carried out to 
evaluate generalization abilities of APAC.

\subsection{Datasets}

Two datasets are used in the experiments.

{\bf MNIST} \cite{lecun-98h}. 
This dataset contains images of handwritten digits with ground truths.
It has 60K training and 10K testing samples.
There are ten types of digits (0-9).
The images are gray-scaled with $28 \times 28$ size.
Background has no texture.

{\bf CIFAR-10} \cite{KrizhevskyMasterThesis}.
This dataset is for benchmarking the 
coarse-grained generic object classification.
It has 50K training and 10K testing samples.
The labels are: plane, car, bird, cat, deer, dog, frog,
horse, ship, and truck.
The images are colored with $32\times 32$ size.
Foreground objects appear in different poses.
Background differs in each image.

\subsection{Image deformation}

Class-indistinctive deformation learning is carried out in all experiments.
Details of deformation are given below.
Some processed images are shown in Fig.~\ref{fig-deformation}.

{\bf Deformation on MNIST.}
We employed random (1) homography transformation, (2) elastic distortion, and (3) line thickening/thinning.

\noindent
(1) {\it Homography transformation}.
Image is projectively transformed by homography matrix $H$.
The eight elements are assigned as Gaussian random variables:
$H_{11}, H_{22} \sim {\cal N}(1, 0.1^2)$,
$H_{12}, H_{13}, H_{21}, H_{23}, H_{31}, H_{32} \sim 
{\cal N}(0, 0.1^2)$.

\noindent
(2) {\it Elastic distortion}.
We followed Simard \etal \cite{Simard03j.c.:best}, 
except for parameter setting.
We used 6.0 standard deviation for the Gaussian filter,
and 38.0 for $\alpha$, the enlargement factor to the displacement fields.

\noindent
(3) {\it Line thickening/thinning}.
Morphological image dilation or erosion is adopted on 
interpolated images, with probabilities $\frac{1}{4}$ and $\frac{1}{4}$,
respectively.
No line thickening/thinning is done with probability of $\frac{1}{2}$.

{\bf Deformation on CIFAR-10.}
We used the ZCA-whitening~\cite{KrizhevskyMasterThesis} 
followed by random (1) scaling, (2) shifting, (3) elastic distortion and side-flipping with probability of $\frac{1}{2}$.

\noindent
(1) {\it Scaling}.
Image is magnified by a factor $s$, randomly picked from continuous uniform distribution, ${\cal U}(1.0, 2.0)$.
Here, $1/s$ is the step size of image interpolation.

\noindent
(2) {\it Shifting}.
Random cropping is applied in following fashion.
The $x$-component of the top-left corner of the interpolated patch is determined by 
sampling a value from ${\cal U}(0, S_x (1-1/s))$,
where $S_x$ is the horizontal size of the original image.
Shift along $y$-axis is determined in the same way but the sampling is done independently.

\noindent
(3) {\it Elastic distortion}.
We used 8.0 standard deviation for the Gaussian filter, and 40.0 for $\alpha$.

A few comments on elastic distortion to CIFAR-10 are in order.
Applying elastic distortion could be harmful for 
images of rigid objects such as plane or car, 
but could be beneficial for images of flexible objects
such as cat or dog.
Nevertheless, we applied elastic distortion to all classes in the same random manner based on two thoughts:
1) A class is not likely to be 
altered by elastic distortion even if 
the resultant image looks somehow unnatural, and 
2) Breaking spatial correlation helps avoiding over-fitting.
It is not our intention to state that 
elastic distortion is particularly important for generic object classification.

\begin{figure}[t]
\begin{center}
\begin{tabular}{cccccccc}
\includegraphics[width=0.088\linewidth]{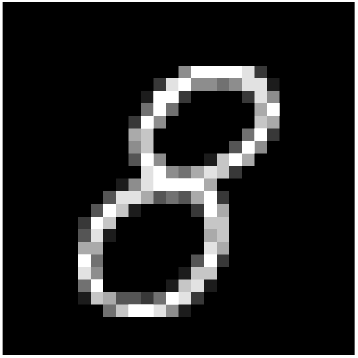} & \hspace{-0.4cm}
\includegraphics[width=0.088\linewidth]{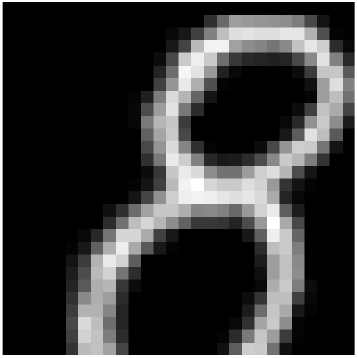} & \hspace{-0.4cm}
\includegraphics[width=0.088\linewidth]{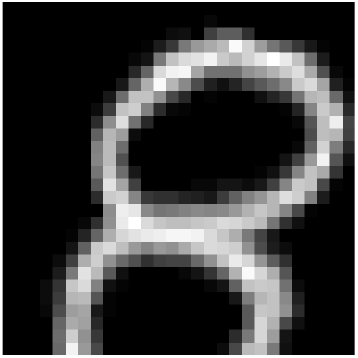} & \hspace{-0.4cm}
\includegraphics[width=0.088\linewidth]{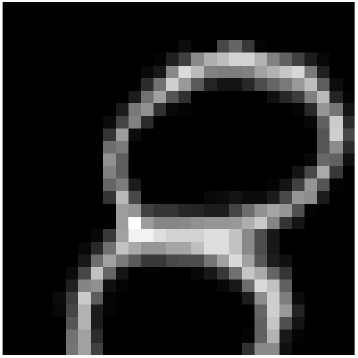} &
\includegraphics[width=0.088\linewidth]{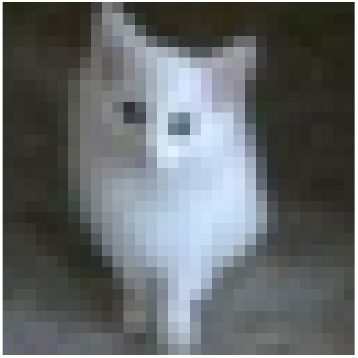} & \hspace{-0.4cm}
\includegraphics[width=0.088\linewidth]{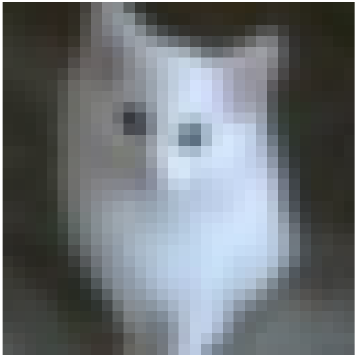} & \hspace{-0.4cm}
\includegraphics[width=0.088\linewidth]{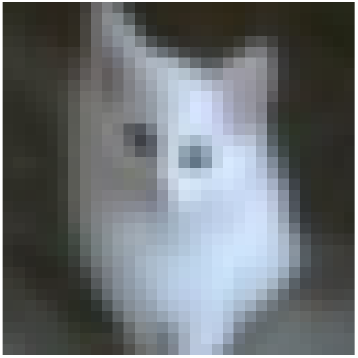} & \hspace{-0.4cm}
\includegraphics[width=0.088\linewidth]{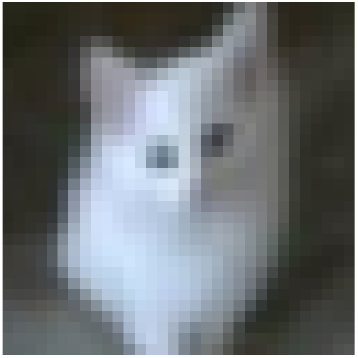}
\\
\includegraphics[width=0.088\linewidth]{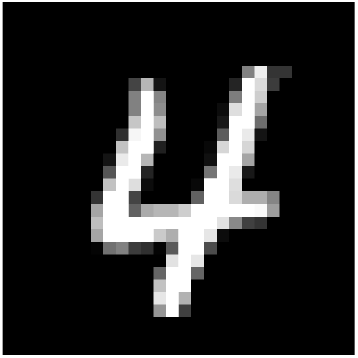} & \hspace{-0.4cm}
\includegraphics[width=0.088\linewidth]{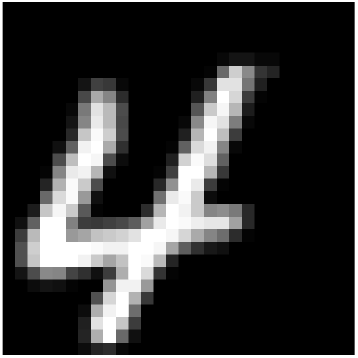} & \hspace{-0.4cm}
\includegraphics[width=0.088\linewidth]{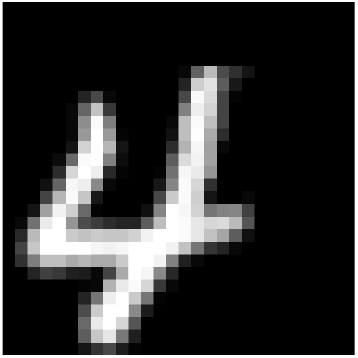} & \hspace{-0.4cm}
\includegraphics[width=0.088\linewidth]{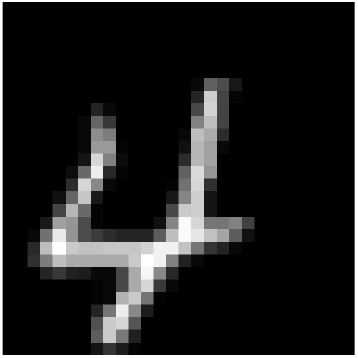} &
\includegraphics[width=0.088\linewidth]{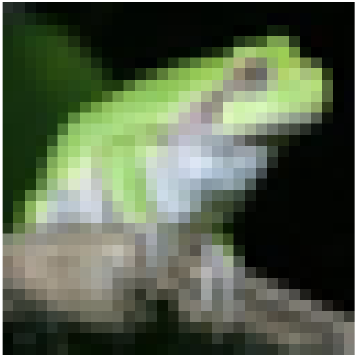} & \hspace{-0.4cm}
\includegraphics[width=0.088\linewidth]{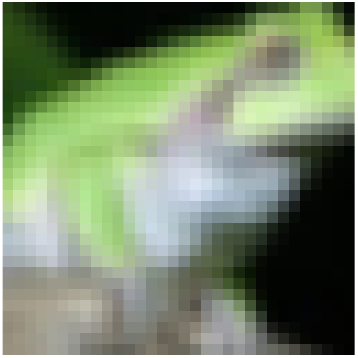} & \hspace{-0.4cm}
\includegraphics[width=0.088\linewidth]{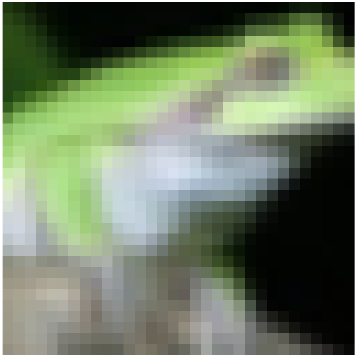} & \hspace{-0.4cm}
\includegraphics[width=0.088\linewidth]{9c_CIFAR10_frog_scale_shift_elastic_08.pdf}
\\
\includegraphics[width=0.088\linewidth]{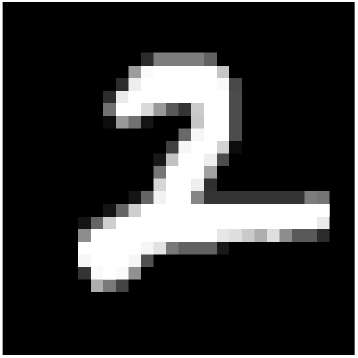} & \hspace{-0.4cm}
\includegraphics[width=0.088\linewidth]{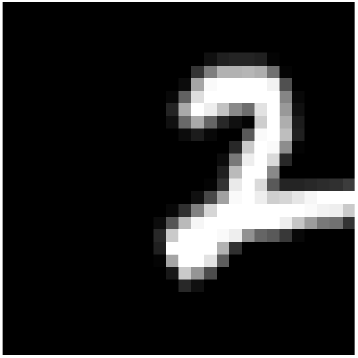} & \hspace{-0.4cm}
\includegraphics[width=0.088\linewidth]{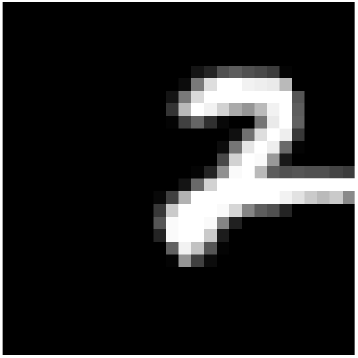} & \hspace{-0.4cm}
\includegraphics[width=0.088\linewidth]{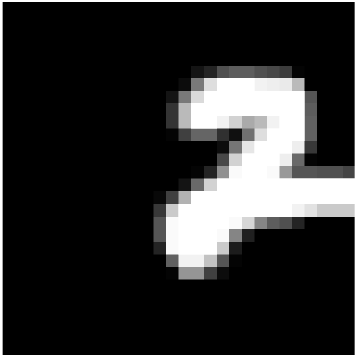} &
\includegraphics[width=0.088\linewidth]{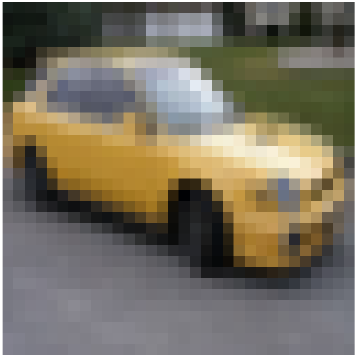} & \hspace{-0.4cm}
\includegraphics[width=0.088\linewidth]{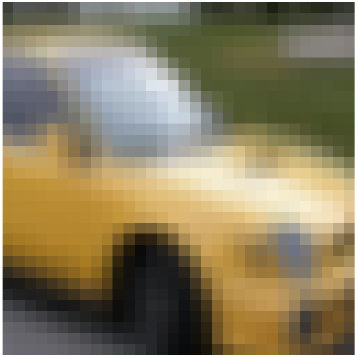} & \hspace{-0.4cm}
\includegraphics[width=0.088\linewidth]{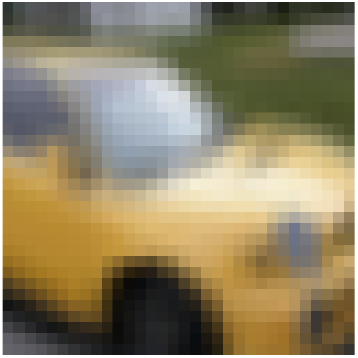} & \hspace{-0.4cm}
\includegraphics[width=0.088\linewidth]{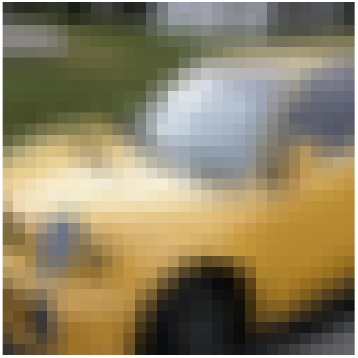}
\\
\includegraphics[width=0.088\linewidth]{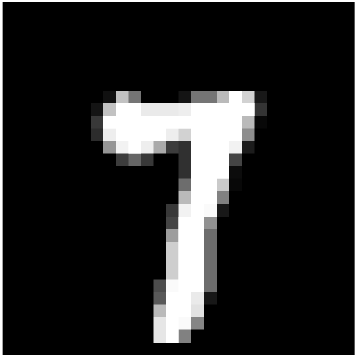} & \hspace{-0.4cm}
\includegraphics[width=0.088\linewidth]{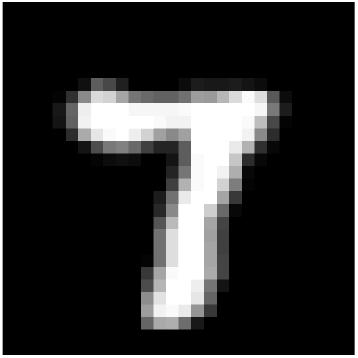} & \hspace{-0.4cm}
\includegraphics[width=0.088\linewidth]{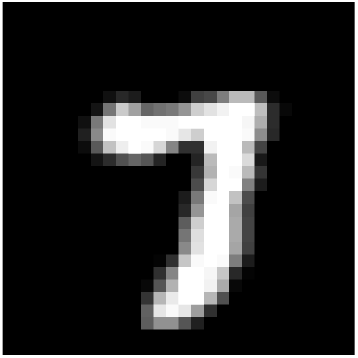} & \hspace{-0.4cm}
\includegraphics[width=0.088\linewidth]{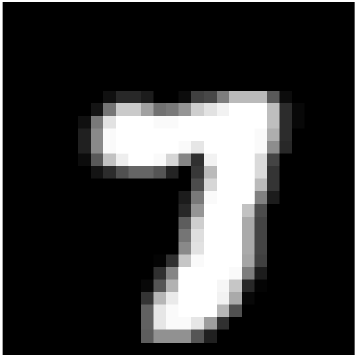} &
\includegraphics[width=0.088\linewidth]{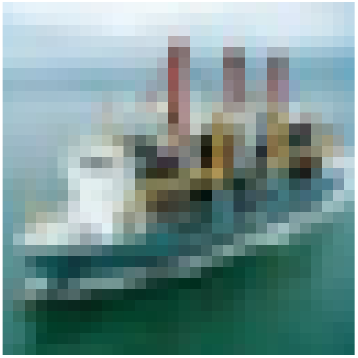} & \hspace{-0.4cm}
\includegraphics[width=0.088\linewidth]{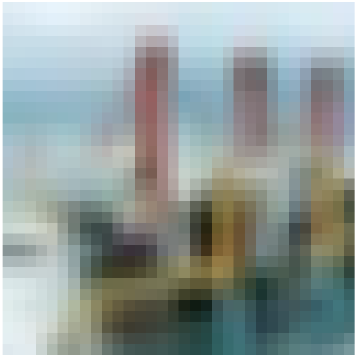} & \hspace{-0.4cm}
\includegraphics[width=0.088\linewidth]{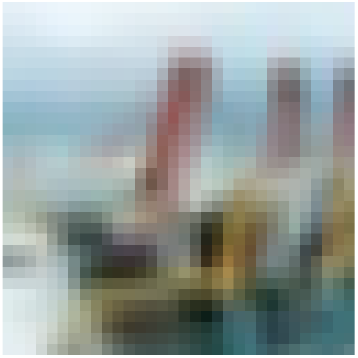} & \hspace{-0.4cm}
\includegraphics[width=0.088\linewidth]{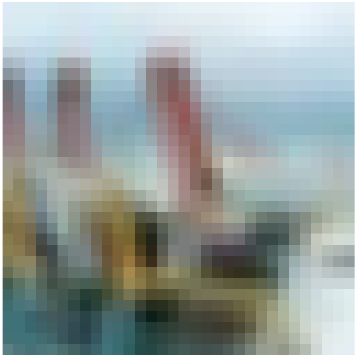}
\end{tabular}
\vspace{-0.5cm}
\includegraphics[width=0.90\linewidth]{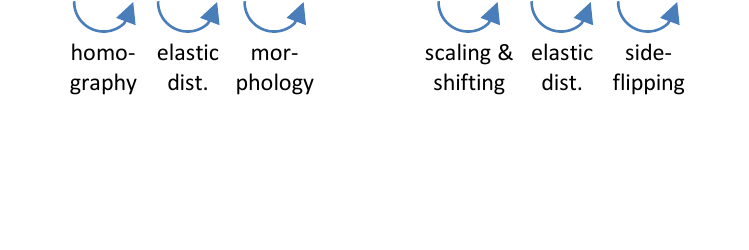}
\end{center}
\vspace{-1.5cm}
{\small \hspace{1.3cm} (a) MNIST \hspace{2.4cm} (b) CIFAR-10}
\caption{Visualization of image deformation. The ZCA-whitening part is skipped for visibility.}
\label{fig-deformation}
\end{figure}

\subsection{Network architectures}

We evaluated CNN and MLP for each of the datasets.
In all networks, the ReLU activation function \cite{NIPS2012_4824} was used.
We trained and evaluated a single model for each of the 
experiments; \ie, no ensembles of classifiers are used.
We did not impose any stochasticity to
the networks during training, such as dropout \cite{2012arXiv1207.0580H}
or dropconnect \cite{icml2013_wan13}.
Network architectures used in our experiments are presented in Table~\ref{table-arch}.

{\bf CNN models.}
For MNIST, 
we used the same numbers of layers and maps in each layer
as in \cite{Ciresan:2012b},
but we used $5 \times 5$ convolutional kernels in all
convolutional layers 
(we use symbol $\mathrm{C}_5$),
whereas different sizes were used in \cite{Ciresan:2012b}.
For CIFAR-10, we just set the architecture by hand 
without any validation.
A non-overlapping maximum pooling with $g \times g$ grid size
(we use the symbol $\mathrm{P}_g$) 
follows each convolution and activation.
We use the symbols F and S for fully-connected and 
softmax, respectively, in the tables.

\begin{table}[t]
\caption{The network architectures. Top: MNIST-CNN model. Middle: CIFAR-10-CNN model. Bottom: MLP models.}
\begin{center}
{\small
\begin{tabular}{|c|ccccccc|}
\hline
layer     & 0 & 1 & 2 & 3 & 4 & 5 & 6 \\
\hline
\# maps   & 1 & 20 & 20 & 40 & 40 & 150 & 10 \\
map size  & $28^2$ & $24^2$ & $12^2$ & $8^2$ & $4^2$ & $1^2$ & $1^2$ \\
operation & $\mathrm{C}_5$ & $\mathrm{P}_2$ & $\mathrm{C}_5$ & 
$\mathrm{P}_2$ & F & F & S \\
\hline
\end{tabular}
\newline
\vspace{0.1cm}
\newline
\begin{tabular}{|c|ccccccccc|}
\hline
layer \h{-0.17cm} & 0 \h{-0.17cm} & 1 \h{-0.17cm} & 2 \h{-0.17cm} & 3 \h{-0.17cm} & \h{-0.17cm} 4 \h{-0.17cm} & \h{-0.17cm} 5 \h{-0.17cm} & \h{-0.17cm} 6 \h{-0.17cm} & \h{-0.17cm} 7 \h{-0.17cm} & \h{-0.17cm} 8 \\
\hline
\# maps \h{-0.17cm} & 3 \h{-0.17cm} & 64 \h{-0.17cm} & 64 \h{-0.17cm} & 128 \h{-0.17cm} & \h{-0.17cm} 128 \h{-0.17cm} & \h{-0.17cm} 256 \h{-0.17cm} & \h{-0.17cm} 256 \h{-0.17cm} & \h{-0.17cm} 128 \h{-0.17cm} & \h{-0.17cm} 10 \\
map size \h{-0.17cm} & $32^2$ \h{-0.17cm} & \h{-0.17cm} $30^2$ \h{-0.17cm} & \h{-0.17cm} $10^2$ \h{-0.17cm} 
& \h{-0.17cm} $8^2$ \h{-0.17cm} & \h{-0.17cm} $4^2$ \h{-0.17cm} & \h{-0.17cm} $2^2$ \h{-0.17cm} & \h{-0.17cm} $1^2$ \h{-0.17cm} & \h{-0.17cm} $1^2$ \h{-0.17cm} & \h{-0.17cm} $1^2$ \\
operation \h{-0.17cm} & \h{-0.17cm} $\mathrm{C}_3$ \h{-0.17cm} & \h{-0.17cm} $\mathrm{P}_3$ \h{-0.17cm} & \h{-0.17cm} $\mathrm{C}_3$ \h{-0.17cm} & \h{-0.17cm} $\mathrm{P}_2$ \h{-0.17cm} & \h{-0.17cm} $\mathrm{C}_3$ \h{-0.17cm} & \h{-0.17cm} $\mathrm{P}_2$ \h{-0.17cm} & \h{-0.17cm} F \h{-0.17cm} & \h{-0.17cm} F \h{-0.17cm} & \h{-0.17cm} S \\
\hline
\end{tabular}
\newline
\vspace{0.1cm}
\newline
\begin{tabular}{|c|cccc|}
\hline
layer & 0 & 1 & 2 & 3 \\
\hline
MNIST & 784 & 2500 & 2000 & 10 \\
CIFAR-10 & 3072 & 4096 & 3072 & 10 \\
\hline
\end{tabular}
}
\end{center}
\label{table-arch}
\end{table}

{\bf MLP models.}
Numbers of layers are determined by validations for both
datasets; it turned out that 3 weight-layers were the best in both datasets. 
Numbers of hidden units for MNIST, 2500 and 2000, are the same as in
\cite{Ciresan:2010}.
Numbers of units for CIFAR-10 are set by hand without any validation.
Softmax normalization is applied to the output units.


\subsection{Training details}

Mini-batch SGD with momentum was used in every experiment.
Initial values of learning rates are: 
$2^{-4}$ for MNIST-CNN,
$2^{-5}$ for MNIST-MLP,
$2^{-8}$ for CIFAR-10-CNN, and
$2^{-8}$ for CIFAR-10-MLP.
Learning rate is multiplied by 0.9993 after each epoch\footnote{Here,
an ``epoch" equals the number of iterations needed to process $N$ virtual samples, where $N$ is the number of 
original training data.}.
The momentum rate is fixed to 0.9 during training.
The mini-batch size is 100.
Training data are randomly sampled with replacement,
meaning that the same empirical sample can be sampled 
more than once in the same mini-batch, 
but deformation is done independently.
Training is terminated at 15K epochs.
We confirmed that 15K epochs gave sufficient convergences through validation.
We added $\mathrm{L}_2$ regularization terms 
with 5e-6 factor to the MNIST-MLP cost function and 
with 5e-7 factor to all the rest of the cost functions.

\subsection{Classification performance}

We first compare classification accuracies between the different decision rules.
Table~\ref{table-summary} shows test error rated produced by APAC and non-APAC.
In the experiments, $M$, the number of virtual samples created from a given image at the testing stage
(see Eq.~(\ref{eq-finite-term-approximation})), is varied from 1$(=4^0)$ to 16,384$(=4^7)$.
Our claim is to use as large $M$ as possible to give class prediction,
so the APAC results shown in Table~\ref{table-summary} are those with $M=$16,384.
In all experiments, 
\begin{table}[h]
\caption{Summary of test error rates produced by our experiments. 
Finite-term approximation with $M=$16,384 is taken in the APAC results.
Non-APAC means the conventional way of prediction, in which each original test sample is fed into the network.}
\begin{center}
{\small
\begin{tabular}{|l|c|c|c|c|}
\hline
\multicolumn{2}{|l|}{Trained on} & \multicolumn{2}{|c|}{augmented data} & original data\\
\hline
\multicolumn{2}{|l|}{Tested by}  & APAC & non-APAC & non-APAC\\
\hline\hline
MNIST    & CNN &  {\bf 0.23\%}  & 0.39\%      & 0.69\% \\
         & MLP &  {\bf 0.26\%}  & 0.29\%      & 1.49\% \\
\hline
CIFAR-10 & CNN &  {\bf 10.33\%}  & 20.05\%    & 22.63\% \\
         & MLP &  {\bf 14.07\%}  & 23.20\%    & 55.96\% \\
\hline
\end{tabular}
}
\end{center}
\label{table-summary}
\end{table}
APAC consistently gives superior accuracies compared to non-APAC
--prediction made by feedforwarding the original test samples-- 
albeit they use the same weight trained with augmented data.

\begin{figure}[t]
\begin{center}
\includegraphics[width=1.00\linewidth]{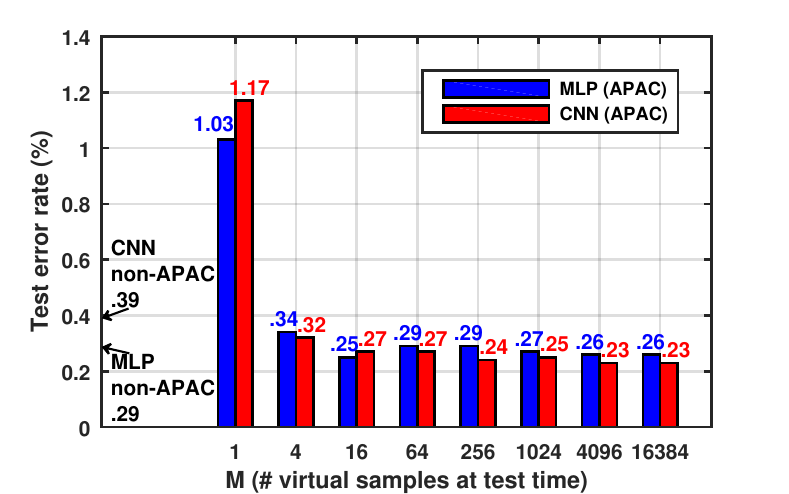}
\end{center}
\caption{Test error rates of our CNN and MLP models on MNIST.
Classification performance of APAC is plotted as a function of the number of virtual samples created at the test time.
Non-APAC (prediction made by a single feedforwarding of an original test sample) results are also shown 
in the figure with texts.
In both cases, the same weights are used.}
\label{fig-test-error-rate-mnist}
\end{figure}

\begin{figure}[tb]
\begin{center}
\begin{tabular}{cccccccccccc}
\hspace{-0.25cm}
\includegraphics[width=0.087\linewidth]{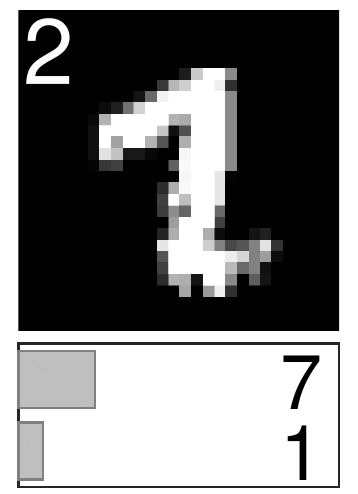} & \hspace{-0.55cm}
\includegraphics[width=0.087\linewidth]{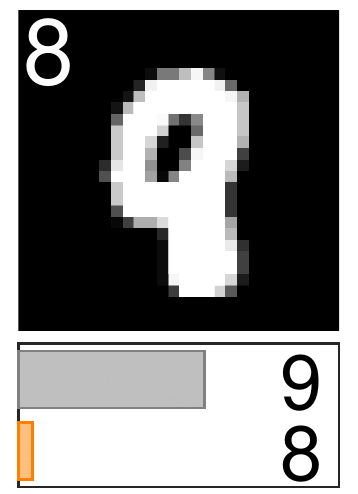} & \hspace{-0.55cm}
\includegraphics[width=0.087\linewidth]{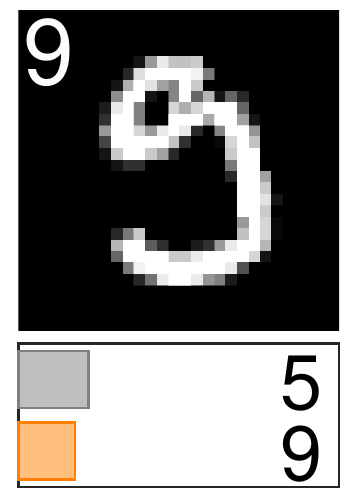} & \hspace{-0.55cm}
\includegraphics[width=0.087\linewidth]{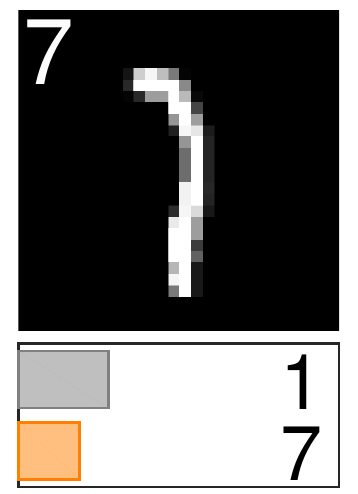} & \hspace{-0.55cm}
\includegraphics[width=0.087\linewidth]{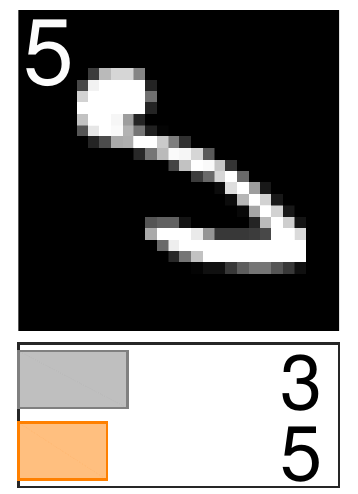} & \hspace{-0.55cm}
\includegraphics[width=0.087\linewidth]{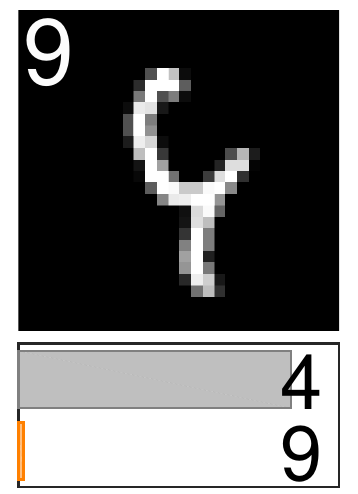} & \hspace{-0.55cm}
\includegraphics[width=0.087\linewidth]{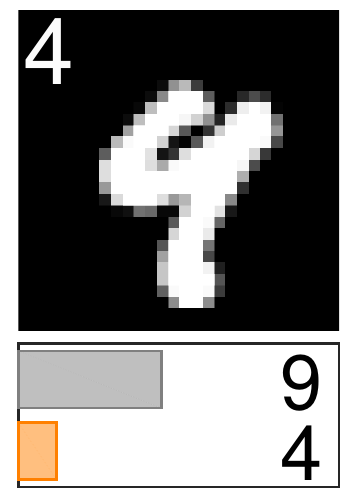} & \hspace{-0.55cm}
\includegraphics[width=0.087\linewidth]{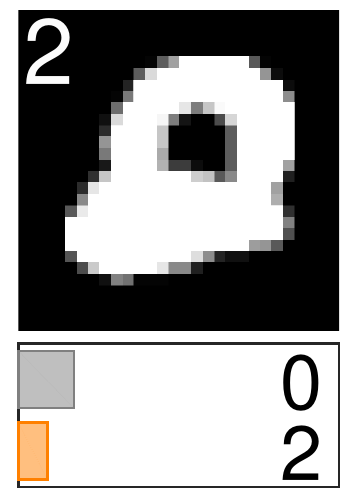} & \hspace{-0.55cm}
\includegraphics[width=0.087\linewidth]{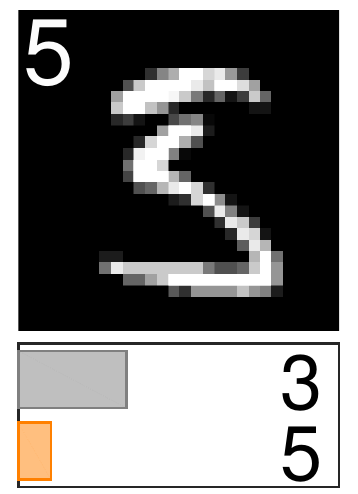} & \hspace{-0.55cm}
\includegraphics[width=0.087\linewidth]{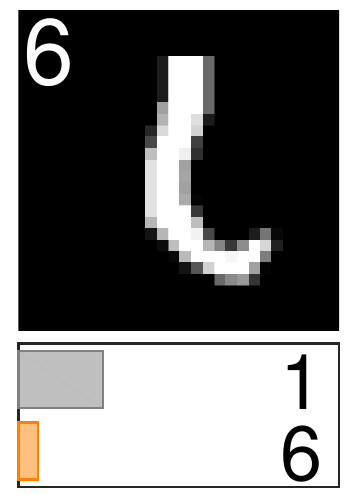} & \hspace{-0.55cm}
\includegraphics[width=0.087\linewidth]{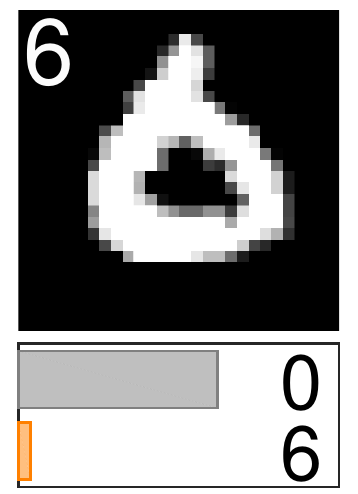} & \\
\hspace{-0.25cm}
\includegraphics[width=0.087\linewidth]{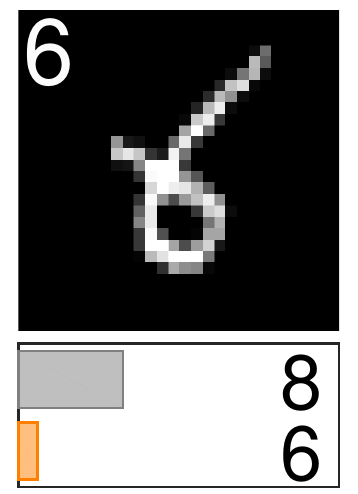} & \hspace{-0.55cm}
\includegraphics[width=0.087\linewidth]{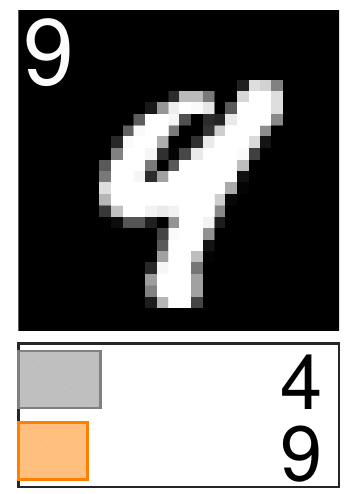} & \hspace{-0.55cm}
\includegraphics[width=0.087\linewidth]{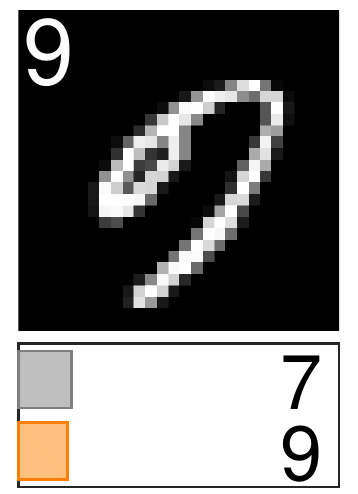} & \hspace{-0.55cm}
\includegraphics[width=0.087\linewidth]{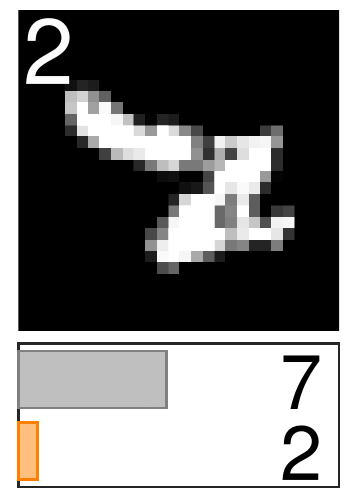} & \hspace{-0.55cm}
\includegraphics[width=0.087\linewidth]{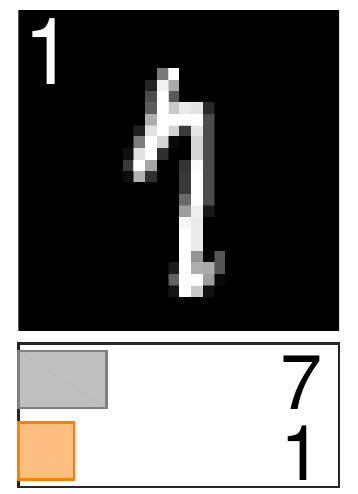} & \hspace{-0.55cm}
\includegraphics[width=0.087\linewidth]{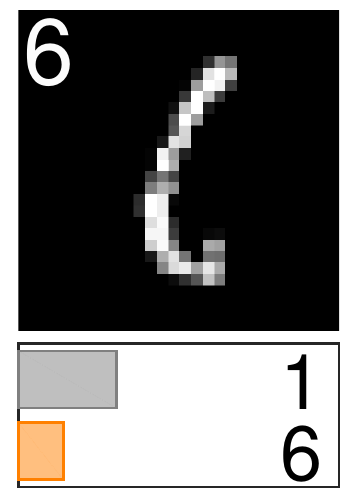} & \hspace{-0.55cm}
\includegraphics[width=0.087\linewidth]{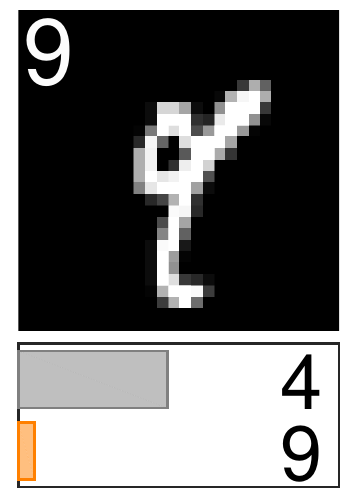} & \hspace{-0.55cm}
\includegraphics[width=0.087\linewidth]{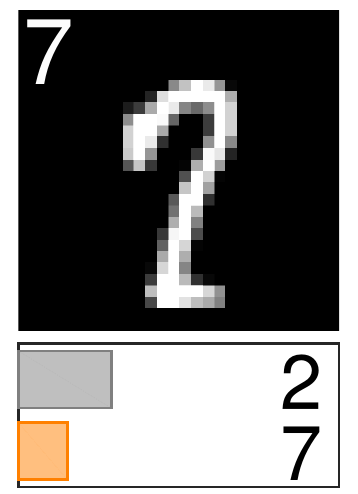} & \hspace{-0.55cm}
\includegraphics[width=0.087\linewidth]{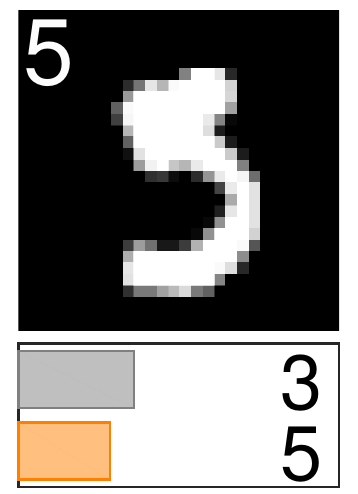} & \hspace{-0.55cm}
\includegraphics[width=0.087\linewidth]{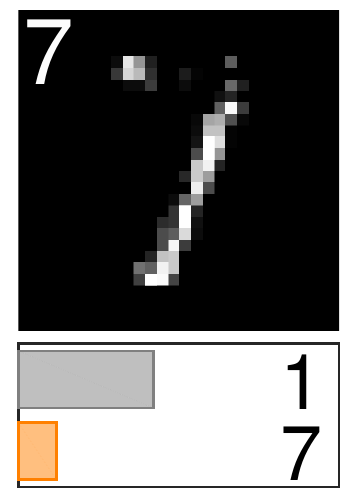} & \hspace{-0.55cm}
\includegraphics[width=0.087\linewidth]{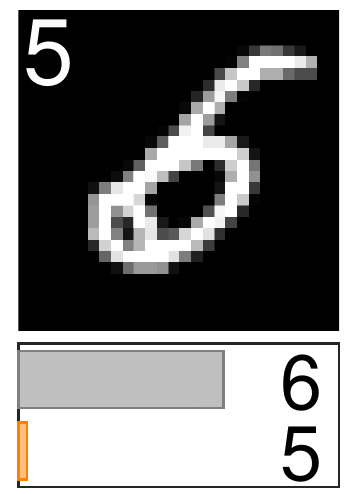} & \hspace{-0.55cm}
\includegraphics[width=0.087\linewidth]{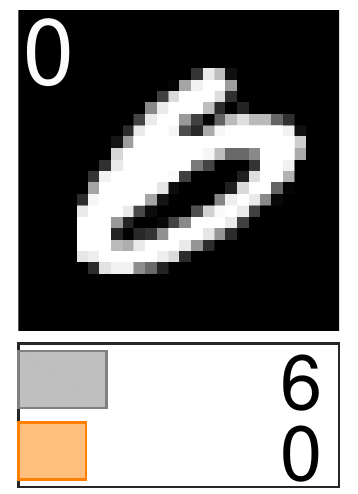}
\end{tabular}
\end{center}
\caption{All MNIST test samples misclassified by our CNN model. 
In each figure, ground truth is printed at the top-left corner.
The bar plot in each figure indicates softmax output of the 1st and 2nd predictions.}
\label{fig-MNIST-misclassified}
\end{figure}

\subsubsection{Performance on MNIST}
We evaluate how classification accuracies change as $M$ goes to a large value.
Plot for $M$ versus the test error rate is shown in Fig.~\ref{fig-test-error-rate-mnist}.
The tendency that the classification accuracy raises as $M$ increases for both networks is clearly observed.
This is due to the fact that the expected loss $J_{\{(x,c)\}}(W)$ is better estimated as $M$ gets larger.

Our CNN model achieved 0.23\% test error rate.
To the best of our knowledge, this test error is the best when a single model is evaluated.
We used no ensemble classifiers, such as model averaging or voting.
(The best test error rate, 0.21\%, was achieved by Wan \etal \cite{icml2013_wan13}, 
where voting of five models was used.)
Training was done only once in each of our experiments.
All misclassified test samples are shown in Fig.~\ref{fig-MNIST-misclassified}.
The top-2 prediction error rate is as low as 0.01\%; \ie, there is 
only one misclassified sample out of 10K test samples, 
with our CNN model.

Our single MLP model achieved 0.26\% test error rate.
To the best of our knowledge, this is the best record among MLP models reported previously.
The best MLP error rate reported in the past was 0.35\%, which was achieved by 
Cire{\c s}an \etal \cite{Ciresan:2010}.
They used a single MLP model that has around 12.1M free parameters and 5 weight layers,
whereas our MLP model has around 7.0M parameters and 3 weight layers.
Though our model is smaller in both the number of free parameters and the network depth, 
ours reaches significantly better classification performance.
Our MLP model has, again, 0.01\% top-2 prediction error rate on the test dataset; \ie,
there is only one misclassified sample.
Interestingly, the very same test sample (shown at the top-left position in Fig.~\ref{fig-MNIST-misclassified}) 
is misclassified by our CNN and MLP models, and all other 9,999 samples are correctly classified within two guesses.

\subsubsection{Performance on CIFAR-10}
\begin{figure}[t]
\begin{center}
\includegraphics[width=1.00\linewidth]{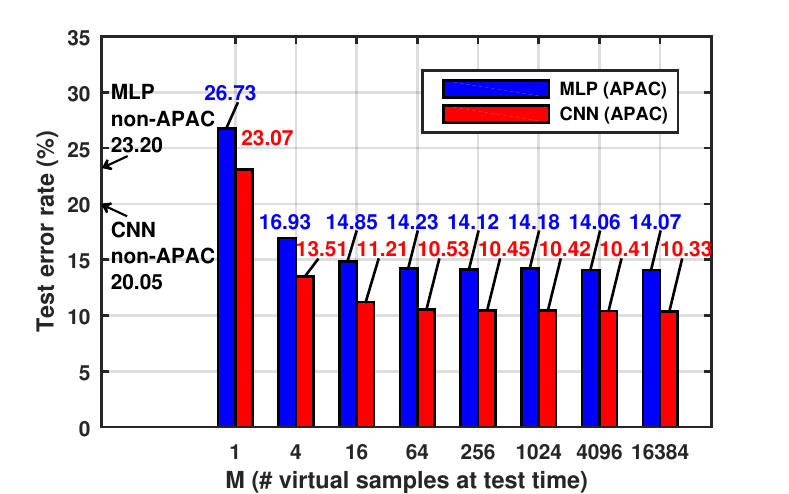}
\end{center}
\caption{Test error rates of our CNN and MLP models on CIFAR-10.
See Fig.~\ref{fig-test-error-rate-mnist} for detail.}
\label{fig-test-error-rate-cifar}
\end{figure}

Plot for $M$, the number of virtual samples generated at the testing stage, 
versus the test error rates is given in the Fig.~\ref{fig-test-error-rate-cifar}.
The tendency that generalization performance raises 
as the number of virtual samples increases is also observed.
Generalization of non-APAC is significantly inferior to that of APAC for both architectures.

Our single CNN model results in 10.33\% test error rate.
This error rate is better than the multi-column CNN (11.21\%)~\cite{Ciresan:2012b} and
the deep CNN reported by Krizhevsky \etal (11\%)~\cite{NIPS2012_4824}, and
worse than the Bayesian optimization method (9.5\%)~\cite{68},
Probabilistic Maxout (9.39\%)~\cite{SprRied2014a}, 
Maxout (9.35\%)~\cite{DBLP:conf/icml/GoodfellowWMCB13}, 
DropConnect (9.32\%)~\cite{icml2013_wan13}, 
Network-in-Network (8.8\%)~\cite{Lin2014}, and
Deeply-Supervised Nets (8.22\%)~\cite{2014arXiv1409.5185L}.
Our result is not close to those of the state-of-the-art methods.
However, we believe that APAC can even improve the generalization abilities of these high-performing methods
if augmented data learning is adopted.

Our single MLP model yields 14.07\% test error rate.
This error rate is worse than the multi-column CNN (11.21\%)~\cite{Ciresan:2012b},
but better than the CNN with stochastic pooling method (15.13\%)~\cite{2013arXiv1301.3557Z} and
the CNN with dropout in final hidden units (15.6\%)~\cite{2012arXiv1207.0580H}.
We are aware that fully-connected neural networks are easy to over-fit when used for image classification tasks.
But still, this experiment gives an evidence that 
a fully-connected network trained with augmented data and tested with APAC can outperform
CNNs trained with recently invented regularization techniques and 
without augmented data~\cite{2013arXiv1301.3557Z, 2012arXiv1207.0580H}.

\subsection{Analysis}

\begin{figure}[t]
\begin{center}
\begin{tabular}{cc}
\h{-0.4cm}
\includegraphics[width=0.54\linewidth]{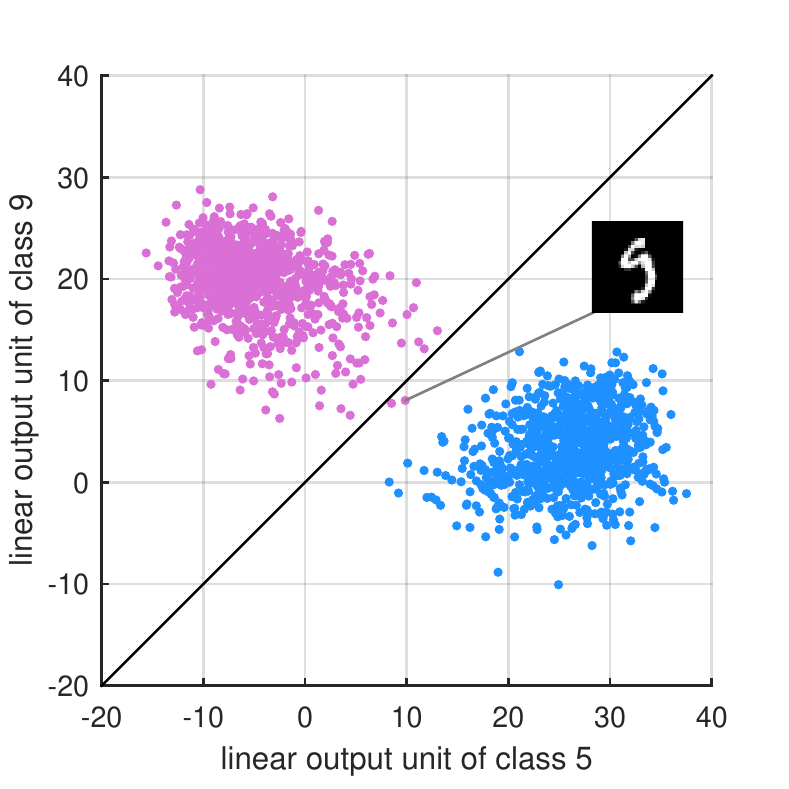} & \h{-0.8cm}
\includegraphics[width=0.54\linewidth]{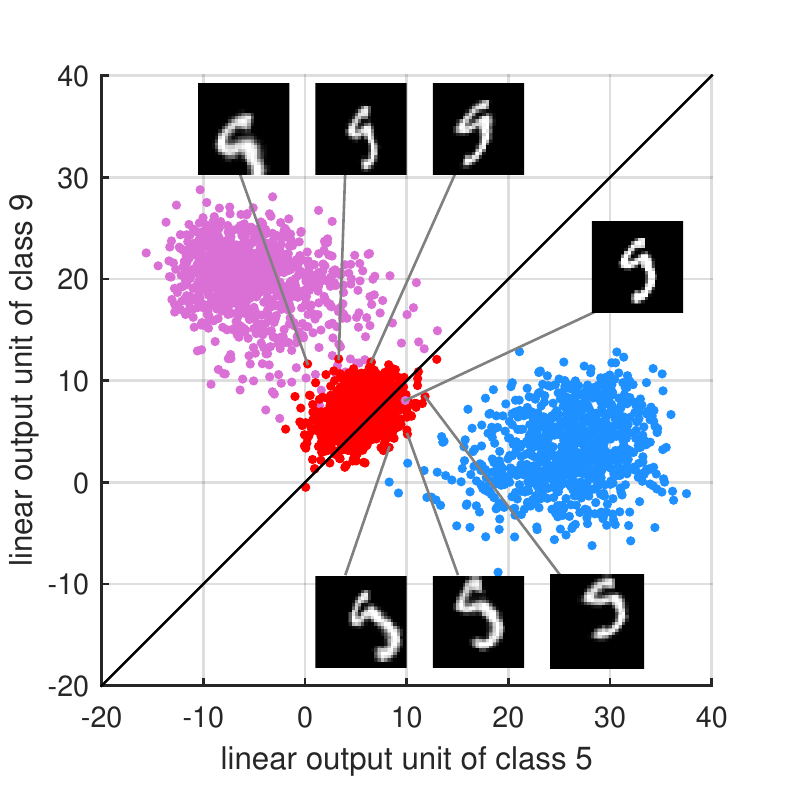} \\
(a) & \h{-0.5cm} (b) 
\end{tabular}
\end{center}
\caption{Illustration of APAC prediction of a class-marginal sample. 
The violet and light blue points corresponds to the class-9 and class-5 test data points, respectively, of MNIST.
The red points corresponds to the virtual data points created from a particular test sample.
See the text for more details.}
\label{fig-linearoutput}
\end{figure}

\begin{figure*}[t]
\begin{center}
\begin{tabular}{cccc}
\includegraphics[width=0.24\linewidth]{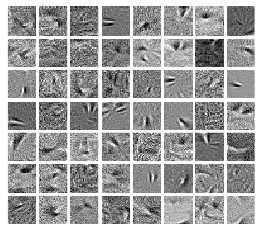} & \h{-0.5cm}
\includegraphics[width=0.24\linewidth]{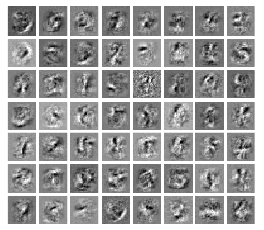} & \h{-0.5cm}
\includegraphics[width=0.24\linewidth]{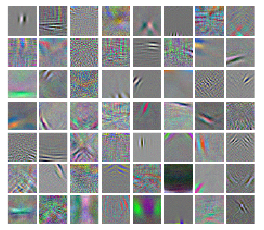} & \h{-0.5cm}
\includegraphics[width=0.24\linewidth]{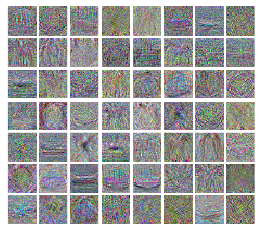} 
\\
{\small (a) augmented MNIST} & \h{-0.5cm}
{\small (b) non-augmented MNIST} & \h{-0.5cm}
{\small (c) augmented CIFAR-10} & \h{-0.5cm}
{\small (d) non-augmented CIFAR-10}
\end{tabular}
\end{center}
\caption{Visualization of randomly selected weight maps in the 1st weight layers of the MLP models
trained with: (a) augmented MNIST, (b) non-augmented MNIST, (c) augmented CIFAR-10, and (d) non-augmented CIFAR-10.}
\label{fig-weightmap}
\end{figure*}

All the experiments we conducted showed that APAC consistently gives better test error rate than non-APAC, 
a way of class prediction through single feedforwarding of original (non-deformed) data, 
when augmented data are learned.
Let us illustrate how the class prediction gets altered between the two decision rules in the case of MNIST.
Figure~\ref{fig-linearoutput}~(a) 
shows a scatter plot of test data points of class-5 and class-9 in a 2D subspace of the linear output space,
with $x$ and $y$-axis corresponding to class-5 unit and class-9 unit.
There, weights are obtained through the class-indistinctive deformation learning, and
plotted data points do not involve image deformation.
A test sample, whose image is superposed in the plot, would be misclassified to class-5 by non-APAC.
We deform this test sample in 1,000 different ways, and plot these virtual data points
in Fig.~\ref{fig-linearoutput}~(b).
One observation is that the virtual data points lie close to the original point.
This is not so surprising because the original and the virtual images share many features in common, and
the network is trained to be insensitive to the differences amongst these samples; namely,
weak homography relation, elastic distortion, and line thickness.
The other observation is that the majority (661 out of 1,000) 
of such virtual data points are in favor of the true class~('9').
Indeed, APAC predicts the true class from the 1,000 virtual samples.
An important point is that there is a better chance of predicting the correct class 
by taking the product of softmax output of many virtual samples created from a given test sample,
rather than by using the softmax output of the test sample.

One might wonder what happens if {\it summation}, instead of {\it product}, of softmax output of many virtual samples
is taken at test stage.
Just for the record, we list the results below. 
Test error rates produced by taking the maximum argument of the softmax sum with $M=$16,384 are: 
0.24\% for MNIST-CNN, 0.27\% for MNIST-MLP, 10.42\% for CIFAR-10-CNN, and 14.01\% for CIFAR-10-MLP.
Softmax product gives better performance in all cases except for the CIFAR-10-MLP.
We do not have a clear explanation why one out of four experiments exhibits opposite result,
but it is safer and more meaningful 
to use softmax product so as to maximize the joint probability
among individual class-probabilities of many virtual instances.

\subsection{Some remarks on augmented data learning}

We make some remarks on how augmented data learning make difference in weights.
Figure~\ref{fig-weightmap} shows the trained {\it weight maps}
in our MLP models.\footnote{Here, a {\it weight map} means a row of weight matrix in the 1st weight layer, 
rearranged in the 2D form
to visualize its spatial weighting pattern.}

{\bf Trained weights for MNIST.}
The weight maps obtained through the augmented data learning have local-feature sensitive patterns
(see Fig.~\ref{fig-weightmap} (a)).
It has been argued that local feature extraction plays an important role
in visual recognition.
Combining local features in a certain way gives discriminative information about the entire object.
CNN is one particular way to embody such strategy.
But, MLP is {\it not}, in a sense that local-feature extractor is not built-in.
Nevertheless, it is not impossible to give local-feature extraction ability to an MLP
as Fig.~\ref{fig-weightmap} (a) indicates.
On the contrary, the weight maps obtained through original data learning have only global patterns 
(see Fig.~\ref{fig-weightmap} (b)), implying that over-fitting to the training data takes place.

{\bf Trained weights for CIFAR-10.}
The weight maps obtained through the augmented data learning exhibit two functionalities (see Fig.~\ref{fig-weightmap} (c)):
the gray-scaled, local-edge extractor 
and spatially-spread, color differentiator.
Similar findings have been pointed out by Krizhevsky \etal~\cite{NIPS2012_4824}.
The weight maps obtained through original data learning exhibit no such functionalities
(see Fig.~\ref{fig-weightmap} (d)).
With lacking spatial structure, the generalization is really poor.

\section{Conclusion}

This paper address an issue of optimal decision rule for augmented data learning of neural networks.
On-line data deformation scheme in network training 
leads a minimization of the loss expectation marginalized over deformation-controlling parameters.
It is expected that robustness against intra-class variation can be trained.
Some sort of SGD can reach one of the local minima of such objective function with finite-term approximation.
The claim is that class decision must be made through similar optimization process;
\ie, the expectation value must be minimized for a given test sample.
This demands that a given test sample must be augmented using the same deformation function used in the training,
to compute the loss expectation for each class, if analytical integration is not feasible.

Our experimental results show that the proposed way of classification, APAC, gives far better
generalization abilities than traditional classification rule, which requires a single feedforwarding of 
a given test sample.
Our CNN model achieved the best test error rate (0.23\%) among non-ensemble classifiers on MNIST.
Top-2 prediction using the model yields a test error rate of 0.01\%.
Through augmented data learning, MLP models acquire local-feature extraction functionality,
which is a key of avoiding over-fitting.
Indeed in the CIFAR-10 experiment, our MLP model using APAC outperforms some CNN models trained with
recently-invented regularization techniques.

{\small
\bibliographystyle{ieee}
\bibliography{arXiv201505_Sato}
}

\end{document}